# Order Matters at Fanatics

Recommending sequentially ordered products by LSTM embedded with Word2Vec

Jing Pan[†]
Fanatics Inc.
San Mateo CA USA
ustcpanjing@gmail.com

Weian Sheng
Airbnb Inc.
San Francisco CA USA
shengweian@gmail.com

Santanu Dey[†]
Fanatics Inc.
San Mateo CA USA
sdey@fanatics.com

## ABSTRACT

A unique challenge for e-commerce recommendation is that customers are often interested in products that are more advanced than their already purchased products, but not reversed. The few existing recommender systems modeling unidirectional sequence output a limited number of categories or continuous variables. To model the ordered sequence, we design the first recommendation system that both embed purchased items with Word2Vec, and model the sequence with stateless LSTM RNN. The click-through rate of this recommender system in production outperforms its solely Word2Vec based predecessor. Developed in 2017, it was perhaps the first published real-world application that makes distributed predictions of a single machine trained Keras model on Spark slave nodes at a scale of more than 0.4 million columns per row.

## 1 Related Work

Pioneering academic research on graph neural network proves an unprecedented advantage for embedding the users and items' node information and topological structure, which are essential for recommendation system [2,4]. Yet none of the existing real-world applications using embedding in the recommender system tackles the unique challenge the e-commerce industry faces: the purchased items of a user form a unidirectional sequence. In academia, next bracket recommendation systems utilizing RNN-like models recommend either the next specific item at a much smaller scale (e.g., tens of thousands of items) [1,7], or simply the next category [5]. The latter next category solution in the industry requires another layer of models to fill the next category with specific items. Industrial deployment of such system has yet been reported for two main challenges both related large scales. The first one is that due to large number of items to recommend, the feature matrix of any kind (e.g., one hot key encoding or user-item matrix) will be very sparse and thus hard to learn with the gradient descent algorithm. The second one is that the cardinal of the output items will also be large, and thus poses challenges in production. The solution to the first challenge was inspired by the use of cosine or Euclidean distance of Word2Vec embeddings in recommendation system in the industry and also in our prior version of bought-together based recommendation. Although our current system was developed independent of the Airbnb 2018 KDD best paper, it bears the same intuition on how to use Word2Vec embedding other than merely a distance. Airbnb applied improved Word2Vec embeddings in its search ranking model [3]. Search ranking is comparable to recommendation: search results are like recommended items; search terms are like user past viewed/purchased items; user demographic remains the same. With feature densified by Word2Vec embedding, building next bracket became feasible for a large number of items.

## 2 Current System

We introduce a component of Fanatics recommender system, the Bought Together-RNN (BTRNN, hereafter). It is the world's first production real-world application of recommender systems that (1) is capable of recommending items in a unidirectional sequence, (2) is capable of predicting near half a million items (aka, the columns of output reach more than 0.4 million per row), (3) uses a combination of Word2Vec embedding as input and LSTM RNN modeling for sequences, and (4) is deployed on Apache Spark. The problem we try to solve is that given a user's

[†]Corresponding authors: Dr. Jing Pan, Santanu Dey: jing.pan@ehealth.com, sdey@fanatics.com. Dr. Jing Pan's current affiliation is eHealth Inc., and Dr. Weian Sheng's previous affiliation is Fanatics Inc.

ordered sequence of items bought in time, what is the next item he or she wants to buy.

Denote $\mathbb{U} = \{1,2,...,m\}$ as the set of m users and $\mathbb{I} = \{1,2,...,n\}$ as the set of n items ever purchased at Fanatics. In our case, n equals to about half a million product ids. We have a very small number of product ids in our catalog that have never been purchased and we don't consider them in our model training process. Let T be the time point up to which we have the historical data of purchase. Let

$$\mathbb{O}_u = \{(t_1, i_1), (t_2, i_2), ..., (t_{m_u}, i_{m_u})\} \ (1)$$

be the set of order history data for user $u \in \mathbb{U}$ having mu orders in the data set. Each order $o_j = (t_j, i_j)$ is a tuple; $j \in (1,2,...,m_u)$; tj is the ascending rank order of purchase time of item j; $i_j \in \mathbb{I}$ is the ordered item. In practice, for items ordered at the same time, we randomly assign a unique purchase time for each item that is smaller than the next order's purchase time and larger than the previous order's purchase time and then assign the purchase time rank according to the assigned purchase time. For example, Alice purchased an apple on at 8 am on Jan 1st, 2016 and purchased a puppy at 8 am on Aug 1st, 2016. The purchase time rank for her apple is 1 and purchase time rank for her puppy is 2.

The goal of the modeling is to model the entire ordered sequence of

$$\left(i_{t_1=1}, i_{t_2=2}, ..., i_{t_j-1}, i_{t_j}, i_{t_j+1}, ..., i_{t_{m_u}}\right) \ (2).$$

Specifically, we don't predict the first item $i_1$ any user would buy. Instead, it is given to the model as an input. We get rid of historical data that users purchased only one item. We only model

$$P(i_{t_j} \mid i_{t_1=1}, i_{t_2=2}, ..., i_{t_j-1}) \ (3)$$

for any $2 = t_2 \le t_j \le t_{m_u}$, and any $i_{t_j} \in \mathbb{I}' \subset \mathbb{I}$, in which $\mathbb{I}'$ is a subset of $\mathbb{I}$ and it contains all items ever purchased except those first items purchased by any user.

We choose to model this problem with a 2-layer stateless LSTM RNN. For this RNN to work, we choose to standardize the length of the sequences to 12 items. The length is chosen between the mean and median of the historical data's purchased item sequence lengths for all users. For the same reason, we tried but did not reuse stateful LSTM because the 12-item length sequences from different users may not necessarily share common properties that require long term memory. Our training validation loss from stateful LSTM with same number of layers and nodes was consistently larger than stateless LSTM RNN. For users with $t_{m_u} \ge 12$, one ordered sequence in formula (2) for a given user is trimmed with moving window of length 12 to $(t_{m_u} - 12 + 1)$ sequences as shown in (4).

$$\left(i_{t_1=1}, i_{t_2=2}, ..., i_{t_{12}}\right)$$
$$...$$
$$\left(i_{t_{m_u}-11+1}, i_{t_{m_u}-12+2}, ..., i_{t_{m_u}}\right)$$
$$(4)$$

For users with $2 \le t_{m_u} < 12$, the sequences are aligned with the last order, and the preceding $12 - t_{m_u}$ elements in the sequence were zero padded (more in the paragraph below) as shown in (5) .

$$\left(0, ..., 0, i_{t_1=1}, i_{t_2=2}, ..., i_{t_{m_u}}\right) \ (5)$$

Our problem (3) is rewritten into

$$P(i_{t'_{12}} \mid i_{t'_1}, i_{t'_2}, ..., i_{t'_{11}}) \ (6)$$

for any $i_{t'_{12}} \in \mathbb{I}' \subset \mathbb{I}$, in which $t'_j$ is the rank order of purchase time in the standardized 12 item order sequence. We encode all items in $\mathbb{I}'$ with one-hot-key encoding, thus the label of our RNN model is a sparse and basis vector of purchased item $i_{t'_{12}} = 1$ over the rest of all items =0 in set $\mathbb{I}'$. The size of the set $\mathbb{I}'$ is more than 0.4 million. Please refer to Part 1 of supplement for how to order the item by purchase time in Spark, as collecting to list following an order by doesn't keep the original order in the order by.

For the items in the input sequence ($i_{t'_1}, i_{t'_2}, ..., i_{t'_{11}}$), We embed $i_{t'_j}$ with vector vi, $i \in \mathbb{I}$, for every purchased item in our history data. We consider every embedding vector vi to belong to $\mathbb{R}^d$. The embedding vi, consists of 102 elements. The 1st element is whether this item has a Word2Vec embedding. The 2nd element is the normalized real number value of a product id. One product id value in our system is a unique integer number. We divide this value with the maximum of all product ids' values. This gives us first way of encoding discrete product ids. We didn't do a one-hot encoding of all product ids, due to implementation considerations. Otherwise, the vector for one item will have 0.5 million elements. We had Word2Vec from previous system implementation readily available, and we have enough reason to believe Word2Vec embedding is a good enough alternative to one-hot key embedding. The 3rd to 102nd elements are corresponding 100-element long vector from Word2Vec embedding of purchased item $i \in \mathbb{I}$. The Word2Vec embedding for all products was trained on sequences of each individual user's past viewed products ordered by ascending viewing time, with a window size of 5 and a vector of 100. The rest of the Word2Vec model uses the default setting in "spark.ml.feature.Word2Vec" in spark2.1. We have another process that trains the Word2Vec embedding. In case the product id cannot be found in the Word2Vec embedding, the 3rd to 102nd elements are all zeros. In a perfect word, Word2Vec embedding should be generated more frequently thus the Word2Vec embedding for a product id will always be found. In the less optimal scenario of the absence of Word2Vec embedding, we hope that the 1st and 2nd elements could step into the model and provide some information for the model. In case $2 \le t_{m_u} < 12$ as shown in (5), we fill the entire 102 elements in first $(12 - t_{m_u})$ of vi with zeros.

The implementation of the stateless LSTM RNN model is described in Part 2 (training in Keras) and Part 3 (distributed prediction on Spark from a single machine Keras model) in the Supplement.

We have two primary use cases for this model. First is personalized product recommendation in regular marketing emails. For a user who subscribes to email and has an order history, we use this model and their previously purchased item sequence to recommend the top items they will buy in their immediate next order. We only make predictions for the next order in this use case. We standardize the length of the sequence the same way in our training (11 item input sequence) as described previously. Second is non-personalized product recommendation. For a given seed product, we pad the first 10 items in the sequence with zero. Then put this seed product in the 11th element in the sequence and predict the top items in the immediate next order. The second application is served for both onsite recommendation, trigger email recommendation, and search ranking. A potential third use case could be predicting the purchase items in the next few orders. To predict the items in the 2nd next order, we need to put the 1st order's top predicted item(s) as the 11th element in the input sequence for the 2nd order. Potentially we could create a feed of products for the user by repeating this process. However, we currently do not have this application in production. Figure 1 is a system design just for the BTRNN component of our recommender system.

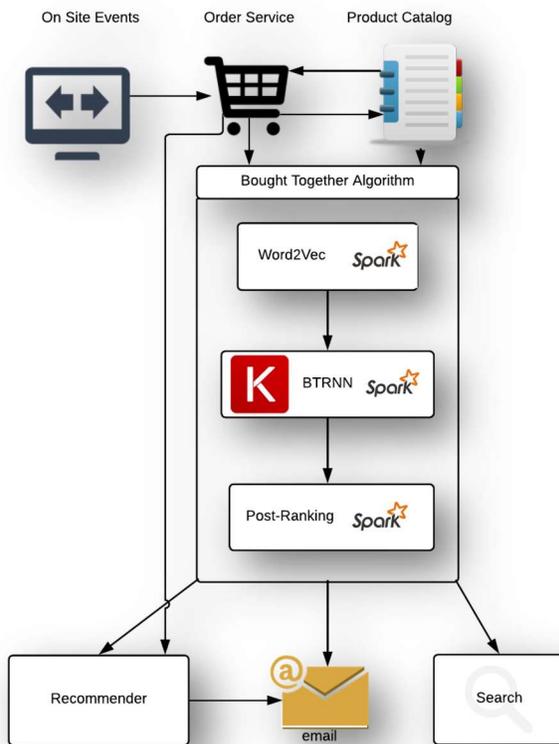

**Figure 1: System Design of BTRNN for Recommender and Other services**

## 3 Results

### 3.1 Predicted Items from Different Algorithms

Figure 2 shows an example of recommendation purely from various algorithms without any additional filtering or ranking. We show an infant product, an infant pacifier clip, as the seed product to better illustrate the effect of time order. As a matter of fact, infant products are very sparse on our site.

First, we can see that BTRNN figures out the seed product is infant related, so among all the humongous selection of shirts we have, it first picks out 3 "Greatest Dad" imprinted shirts, although not from the exact same team, but all from baseball leagues. Then it picks up a woman's shirt, then a bottle that can be used as a baby bottle, and then a youth sleep set. It shows that RNN model figures out the baby product is strongly related to the term "Greatest Dad". Given that our customer base are more male sports fans than female, it doesn't pick out "Mom" products. It also figures out baby grow up to be youth. Apparently, it places less importance on the team for a relatively sparse baby product, but for grownup men/women's shirt, it will pick shirts from the same team and gender automatically among top predicted products. If this youth sleep set shown in Figure 2 become another seed product, the top recommended products from BTRNN doesn't contain any infant products, but have same team junior products (even older than youth).

Second, Word2Vec, on the other hand, naturally places more emphasis on the same team. It picks various products from the same team. The same team youth t-shirt, in this case, is a good pick, but the problem is that when this youth t-shirt become the seed product, Word2Vec recommends this exact clip at the top rank, which is less optimal because baby only grows up, not the other way around. Third, the basic co-occurrence matrix outputs same team baby products. Those are really good predictions, but they fixate at the seed product's purchase time and lack diversity. Based on the above observations, we completely replaced the output of bought together algorithm from Word2Vec to BTRNN in our production system.

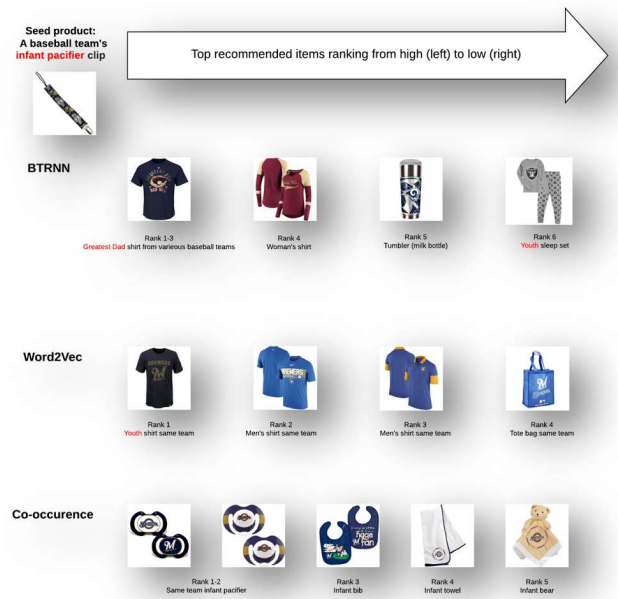

**Figure 2: Example of Recommended Items from BTRNN, Word2Vec and co-Occurrence**

### 3.2 Performance in Emails in Production

In our first application of BTRNN in regular marketing emails, we replaced Word2Vec with BTRNN. The growth rate of the average order value per email for emails with BTRNN compared to emails with Word2Vec is higher than the growth rate of year to year change alone. We have a relatively clean pre-post test of switching bought together algorithm in the second application of trigger emails. In this application, click to open rate increased within the range of 0.5 to 1 percent, which is a significant gain for this type of email which normally has a very high open rate but a very low click rate historically. We did not release specific numbers here due to confidentiality reasons. The price of the recommended items from BTRNN is neither significantly deviate from (a) the price of the seed products of email subscribers who are a subset of all customers nor (b) the price of all purchased items for all customers.

## 3.3 Offline Validation of Actual Immediate Next Ordered Items

BTRNN is performing better than random. In a perfect world, the true immediate item will have a rank of 1 in the BTRNN's output of more than 0.4 million items. In reality, the true immediate next item will be ranked in other positions among the more than 0.4 million items. The smaller the predicted rank of the true immediate item, the better the BTRNN's prediction is. Prior to our more than 0.4 million item model released in production, we trained a smaller model with 137,667 items to predict. The true immediate next item in the testing set on overage ranked 2031(rounded) in our 137,667 possible items. That is, the true item's average rank is among top 2031/137,667=1.5% of all possible predicted items. If BTRNN predict next item completely randomly, the average rank would be 137,667/2=69k, and one can do a Wilcoxon signed-rank test for each product with randomly generated rank.

We don't have an apple to apple comparison to Word2Vec in offline validation, as Word2Vec predicts items in $\mathbb{I}$ while BTRNN predicts items in $\mathbb{I}'$. We only know that when Word2Vec was in production in 2017, any true next bought items (not limited to the immediate next true item) had about 20% chance to be among the top 100 predicted items. We do not have information on the Word2Vec's ranks of the other 80% of true bought items which Word2Vec failed to predict as top 100 recommended items. The reason is that in production, we intentionally did not to score the entire set of $\mathbb{I}$ by a pre-filtering process and we only store top 100 recommended items. The Word2Vec's ranks of the other 80% of true bought items were expected to be very large.

## 4 Discussion and Future Work

The current BTRNN algorithm successfully addresses the use case of the recommendation for an e-Commerce customer often times requiring a unidirectional array of items on the time axis. It is the first recommender system that models with LSTM RNN and at the same time embed with Word2Vec. In 2017, we were perhaps the first team in the world to make distributed predictions from Keras model on Spark cluster's slave nodes, as opposed to on master node only, at the scale of more than 0.4 million columns per row for output.

The BTRNN algorithm in production is friendly to new user recommendation but in its current form is not friendly to new item discovery. The moment a new user show interaction with an existing item (view or buy) called the seed item, the BTRNN algorithm can generate recommended items. There are two types of new items for the cold start problem. On one hand, if the new item is the seed item, we simply use another algorithm to do recommendation, until the Word2Vec embedding process captures the embedding of the new item after the new item would have been purchased. On the other hand, if the new item is to be recommended, we use similar item recommendation algorithm to find new items similar to BTRNN recommended old items to put new items in the recommended item list.

The BTRNN sequence model in its current form is the right model for expressing the impact of the order in terms of the immediate next order of either a seed item or a sequence of seed items. We use the normalized discounted cumulative gain (NDCG, hereafter) as our golden rule before releasing any of our recommendation model to the production system. One can develop metric for the true and predicted sequences of items in recommendation system, or can adopt one from DNA sequencing.

Our BTRNN although in the current form is not yet a graph model, but we hope it will inspire further work on graph neural network. Firstly, the items are like nodes in graph. Word2Vec preserves the linear substructure of the items from the bought together user-item matrix. Instead of a directionless distance in Word2Vec embedding or user-item matrix, the LSTM model further adds directional edges to the items with possible edge value of LSTM predicted next bought probability. Secondly, the success of BTRNN algorithm provides a universal paradigm for stacking deep learning model on top of any other real graph-based embedding. Thirdly, our engineering solution to put a Keras model on Spark is brutal force such that we give predictions on the entire item set. Graph partitioning can potentially reduce the amount of items to be predicted, and the same solution can improve Tencent's (Chinese social media) common friends discovery algorithm which also used brutal force solution on Spark [6].

The BTRNN algorithm has attracted broad interests in the industry. At eHealth Inc., sales calls can be decomposed into conversational elements. Modeling the sequence of conversational elements from top agents with a graph or BTRNN, and extracting common paths can help gain sales insights at scale. Potential applications of such model include but not limited to pre-job agent training programs or real time agent sales coaching system (Varghese and Zhou, personal communication).

Supplement

1. Order items by time stamp in Spark

On Spark cluster (regardless your language choice), ordering items by their time stamp doesn't keep the order across nodes when followed by collecting to a list of items, regardless of whether collect by key (such as UUID) or not. Aka, any variation of df.orderby.tolist doesn't work. Please refer to this stack overflow discussion [1] for a description of the problem. We did it the following way:

```
1 %pyspark
2 import operator
3 def sorter(l):
4     res = sorted(l, key=operator.itemgetter(0))
5 sqlContext.udf.register("sort_udf", sorter)
6
7 def zip_df_by_sorted_list\
8 (df_in, group_by_list,order_by_col,value_col,order_list_col,value_list_col):
9 '''
10 df_in: key_id (eg, UUID),order_col (eg, time stamp),value_col (eg, product_id)
11 df_out: key_id,[order_key_by_order_col],[value_order_by_order_col]
12 '''
13    group_by_str=''
14    for i in group_by_list:
15        group_by_str=group_by_str+i+','
16    group_by_str=group_by_str[:-1]
17
18    df_in.registerTempTable('df_in')
19    df_grouped=sqlContext.sql('''
20    select {0},collect_list(struct({1},{2})) as value_struct_list,
21    collect_list(struct({1},{1})) as order_struct_list
22    from df_in
23    group by {0}'''.format(group_by_str,order_by_col,value_col))
24
25    df_grouped.registerTempTable('df_grouped')
26
27    df_out=sqlContext.sql('''
28    select a.*,sort_udf(order_struct_list) as {0},
29    sort_udf(value_struct_list) as {1}
30    from df_grouped a
31    '''.format(order_list_col,value_list_col))
32    return (df_out)
33
34 df_out=zip_df_by_sorted_list(df_in,\
35 group_by_list,order_by_col,value_col,order_list_col,value_list_col)
```

2. LSTM model Training

The structure of the model itself was extremely simple done in Keras 1.0 or later, on AWS P2.8xlarge machine with tensorflow_gpu backend. We used a fit_generator giving the amount of data we have.

```python
#python
1 model = Sequential()
2 model.add(LSTM(600, input_shape=(seq_len, feature_len), return_sequences=True))
3 model.add(LSTM(600))
4 model.add(Dense(n_index, activation='softmax'))
5 model.compile(loss='categorical_crossentropy', optimizer='adam')
```

3. Predicting Keras model on Spark slave nodes

The key is that the model object has to be loaded as a global variable. The actual model hdf5 file needs to be stored on a s3, or HDF *etc.* location where your slaves can access. Then you download your model file to your slave node. You have to add a uuid to the model file's local path. The local path needs to have enough disk space. If you are on AWS, make the local path on an additionally mounted EBS volume. There are more than one downloaded model files on the disk of one slave. Then load the local model file to your model object as a global variable. When making predictions, call the model object from any map procedure. In this case, we used the withColumn, but feel free to choose your own.

```
1 %pyspark
2 spark=sparkSession("start your spark session, set enable your hive context to true")
3 df_in=spark...#create your input data frame however you want
4 __MODEL__ = None
5 def predict (one_row_of_data,bucket,key, other_parameters):
6   global __MODEL__
7   if __MODEL__ is None:
8     local_file_path = \
9 "common_file_path_on _EBS"+str(uuid.uuid1())+".file_extension"
10    s3 = boto3.resource('s3')
11    s3.Bucket(bucket).download_file(key, local_file_path)
12    __MODEL__=load_model(local_file_path)
13 prediction=__MODEL__.predict(one_row_of_data)
14   output=whatever_you_want_to_do_with_prediction(prediction, other_parameters)
15   return (output)
16 predict_udf=udf(predict, whatever_return_type_you_need)
17 df_out=df_in.withColumn("out_put",predict_udf(df_in.input))
```